\documentclass{article}

\usepackage[nonatbib,preprint]{neurips_2024}

\usepackage[utf8]{inputenc} % allow utf-8 input
\usepackage[T1]{fontenc}    % use 8-bit T1 fonts
\usepackage{hyperref}       % hyperlinks
\usepackage{amsfonts}       % blackboard math symbols
\usepackage{nicefrac}       % compact symbols for 1/2, etc.
\usepackage{microtype}      % microtypography
\usepackage[dvipsnames,table]{xcolor}
\usepackage{graphicx}
\usepackage{microtype}
\usepackage{hyperref}
\usepackage{url}
\usepackage{booktabs}
\usepackage{subcaption} % 引入子图包
\usepackage{multirow} % For multirow
\usepackage{array}       % For centering columns
\usepackage{pifont}
\usepackage{wrapfig}
\usepackage{marvosym}
\usepackage{amsmath,amsfonts,bm}

\def\eqref#1{equation~\ref{#1}}
\def\1{\bm{1}}

\def\vt{{\bm{t}}}

\def\vx{{\bm{x}}}

\def\vz{{\bm{z}}}

\DeclareMathAlphabet{\mathsfit}{\encodingdefault}{\sfdefault}{m}{sl}
\SetMathAlphabet{\mathsfit}{bold}{\encodingdefault}{\sfdefault}{bx}{n}

\title{ConvLLaVA: Hierarchical Backbones as Visual Encoder for Large Multimodal Models\thanks{\textsuperscript{\Letter} Corresponding author.}}

\renewcommand\footnotemark{}

\author{
\!\!Chunjiang Ge$^1$, Sijie Cheng$^3$, Ziming Wang$^2$, Jiale Yuan$^2$, Yuan Gao$^2$\\\textbf{Jun Song}$^2$\textbf{,} \textbf{Shiji Song}$^1$\textbf{,} \textbf{Gao Huang}$^{1,}$\textsuperscript{\Letter}\textbf{,} \textbf{Bo Zheng}$^2$ \\
$^1$Department of Automation, Tsinghua University \\
$^2$Alibaba Group\\
$^3$Department of Computer Science and Technology, Tsinghua University\\
\texttt{gecj20@mails.tsinghua.edu.cn}
}

\begin{document}
\maketitle

\begin{abstract}
High-resolution Large Multimodal Models~(LMMs) encounter the challenges of excessive visual tokens and quadratic visual complexity. Current high-resolution LMMs address the quadratic complexity while still generating excessive visual tokens. However, the redundancy in visual tokens is the key problem as it leads to more substantial compute. To mitigate this issue, we propose ConvLLaVA, which employs ConvNeXt, a hierarchical backbone, as the visual encoder of LMM to replace Vision Transformer~(ViT). ConvLLaVA compresses high-resolution images into information-rich visual features, effectively preventing the generation of excessive visual tokens. To enhance the capabilities of ConvLLaVA, we propose two critical optimizations. Since the low-resolution pretrained ConvNeXt underperforms when directly applied on high resolution, we update it to bridge the gap. Moreover, since ConvNeXt's original compression ratio is inadequate for much higher resolution inputs, we train a successive stage to further compress the visual tokens, thereby reducing redundancy. These optimizations enable ConvLLaVA to support inputs of 1536$\times$1536 resolution generating only 576 visual tokens, capable of handling images of arbitrary aspect ratios. Experimental results demonstrate that our method achieves competitive performance with state-of-the-art models on mainstream benchmarks. The ConvLLaVA model series are publicly available at \url{https://github.com/alibaba/conv-llava}.
\end{abstract}

\section{Introduction}

Large Multimodal Models (LMMs;~\cite{gpt4v,gemini,claude3}) have achieved notable advancements in recent years, demonstrating superior performance in diverse domains, including image and video understanding~\cite{ureader,xc2-4k}, digital agent development~\cite{appagent}, and robotics~\cite{roboflamingo}. The imperative to comprehend a wide range of tasks and intricate scenes underscores the critical role of the visual encoder, which is mostly a Vision Transformer~(ViT;~\cite{vit}). However, ViT's quadratic spatial complexity and output of excessive visual tokens limit its application in diverse and high-resolution tasks~\cite{ureader,li2023otterhd,xc2-4k, cheng2023can}. The excessive visual tokens lead to a significant computational burden in the Large Language Model~(LLM;~\cite{llama,llama2}), far exceeding the computational cost imposed by the quadratic spatial complexity in the visual encoder. Such redundancy in the visual tokens not only sacrifices efficiency but also impedes the effective extraction of visual information~\cite{llava-v1-6,xc2-4k}. While a range of methods~(Tab.~\ref{tab:table-1};~\cite{llava-v1-6,li2023monkey,vary}) have been proposed to remedy the quadratic spatial complexity of ViT, they fail to mitigate the key problem, the redundancy in the visual tokens~\cite{fastv,lin2023vila}. 

Hierarchical visual backbones~\cite{resnet,senet,davit}, which can be considered as counterparts to ViT, can well address the problem of excessive visual tokens due to their inherent \textit{\textbf{Information Compression}} process. Specifically, features are sequentially compressed across stages in hierarchical backbones. They compress visual features by \textit{32$\times$}~\cite{resnet,liu2022convnet} compared to ViT with only \textit{14$\times$}~\cite{vit}. Therefore, at the same resolution they generate fewer than \textit{1/4} visual tokens compared to ViT, significantly alleviating computational burdens on the LLM. Moreover, hierarchical visual encoders, typically designed with linear spatial complexity~\cite{liu2022convnet,davit,resnet}, effectively tackle both the issue of excessive visual tokens and the quadratic visual complexity.

\begin{table}[!t]
  \caption{Comparison with previous methods. Res., VE, \#V Tokens denote resolution, visual encoder, and the number of visual Tokens. Enumerate aspect ratio (Enum) indicates that the model supports a set of predefined aspect ratios. Fix aspect ratio means the model supports fixed resolution input. Any aspect ratio means the model supports arbitrary aspect ratio input. $^*$: OtterHD does not actually have a visual encoder. The spatial complexity for its visual tokens is quadratic. }
  \label{tab:table-1}
  \centering
    \renewcommand{\arraystretch}{1.1}
    \scalebox{0.95}{
    \tabcolsep3pt
    \begin{tabular}{lccccccc}
    \toprule
    \multirow{2}{*}{Method} & \multirow{2}{*}{Res.} & \multirow{2}{*}{\#V Tokens} & \multicolumn{2}{c}{Complex Design} & \multicolumn{3}{c}{Visual Encoder}           \\
                        &                             &                                & Cropping         & Extra VE        & Model            & Complexity & Aspect Ratio \\ \midrule
    LLaVA-1.5               & 336                         & 576                            &                  &                 & ViT            & Quadratic  & Fix          \\
    OtterHD                 & 1024                        & 1225                           &                  &                 & None             & Quadratic$^*$  & Any          \\
    LLaVA-NExT              & 672                         & 2880                           &    \textcolor{red}{\ding{51}}              &                 & ViT            & Linear     & Enum         \\
    MiniGemini-HD           & 1536                        & 2880                           &      \textcolor{red}{\ding{51}}            &       \textcolor{red}{\ding{51}}          & ViT,ConvNeXt & Linear     & Enum         \\
    \midrule
    \rowcolor{cyan!15}ConvLLaVA          & 1024                        & 256                            &                  &                 & ConvNeXt       & Linear     & Any          \\
    \rowcolor{cyan!15}ConvLLaVA          & 1536                        & 576                            &                  &                 & ConvNeXt       & Linear     & Any      \\                       
    \bottomrule
    \end{tabular}}
\end{table}

We choose to employ ConvNeXt among the hierarchical visual encoders due to its excellent performance~\cite{convnext-vs-vit,fc-clip} and the availability of off-the-shelf contrastive language-image pretrained weights~(CLIP;~\cite{clip}), which mainstream visual encoders of LMMs adopt~\cite{blip2,llava-v1,qwen-vl,mm1}. However, directly replacing ViT with ConvNeXt leads to inferior performance on general capabilities benchmarks~(Section~\ref{sec:updating}). This can be attributed to the fact that ConvNeXt is pretrained on low resolution, whereas we directly apply it to high-resolution~\cite{openclip,laion5b}. Moreover, the pretraining data for ConvNeXt is considered to be of low quality~\cite{metaclip,openclip,laion5b} compared to ViT's pretraining data~\cite{clip}. To address these issues, we propose to update the visual encoder rather than freezing it. Surprisingly, updating the visual encoder enables ConvNeXt to perform comparably to ViT on general benchmarks. On fine-grained benchmarks, we observe that ConvNeXt outperforms ViT. These findings indicate that even when compressing visual tokens to an equal quantity, the higher resolution model's features still contain more fine-grained information. This observation inspires us to further scale up the resolution. However, further scaling the resolution beyond 1024 leads to the generation of excessive visual tokens. To mitigate this issue, we further compress the visual information with an additional ConvNeXt stage to enhance the inherent \textit{information compression} of hierarchical backbones. The visual inputs would be compressed by \textit{64$\times$} rather than \textit{32$\times$} to further reduce the redundancy. Hence, ConvLLaVA generates only 576 visual tokens when processing 1536 resolution inputs, which is equivalent to the number of visual tokens generated by ViT when processing 336 resolution inputs~(Section~\ref{sec:add-stage}).

In summary, we introduce ConvLLaVA whose visual encoder is a five-stage ConvNeXt. ConvLLaVA compresses high-resolution images into information-rich visual features, effectively avoiding the generation of excessive visual tokens~(in Tab.~\ref{tab:table-1};~\cite{llava-v1-6,li2023monkey,minigemini,llava-hr}). Furthermore, thanks to the translation equivalence of convolution, ConvLLaVA can be trained on low-resolution and evaluated on higher resolutions, and it can also handle images of arbitrary aspect ratio. Extensive experiments have demonstrated the effectiveness of our method. ConvLLaVA 7B outperforms LLaVA-1.5-13B across various benchmarks, including MME~\cite{mme}, MMBench~\cite{liu2023mmbench}, SEEDBench~\cite{li2023seed}, RealWorldQA~\cite{grok1_5}, TextVQA~\cite{textvqa}, DocVQA~\cite{docvqa}, POPE~\cite{pope}, and MMVet~\cite{mmvet}. 
\section{Related Work}

\textbf{Large Multimodal Models.} To harness the potential of Large Language Models and incorporate visual information, BLIP series models~\cite{blip2,dai2023instructblip} propose the Q-former, which generates visual tokens for LLMs to interpret visual data. Meanwhile, LLaVA~\cite{llava-v1} employs a single linear layer to map visual features to the word embedding space, allowing LLMs to perceive vision features. 
These approaches utilize the ViT as the visual encoder~\cite{clip,vit,honeybee,lin2023vila,minigpt}, primarily tailored for low-resolution visual data (e.g., 224 or 336 resolution). 
Moreover, Qwen-VL~\cite{qwen-vl} and mPLUG-owl2~\cite{mplug-owl2} scale the resolution of ViT to 448 by updating the weights of ViT. However, these methods fail to further scale up resolution due to the quadratic spatial complexity of ViT, while ConvNeXt can scale up the resolution with the linear cost increase. Qwen-VL~\cite{qwen-vl} and mPLUG-owl2~\cite{mplug-owl2} also explore to reduce the visual tokens via resampler. However, recent studies~\cite{honeybee,xc2-4k} show that convolution or simply concatenation performs better than resampler. 

\textbf{High-resolution LMMs with Cropping.} 
The representative cropping method for high-resolution LMMs is introduced in LLaVA-NExT~\cite{llava-v1-6}, which partitions an image into four patches, each encoded separately by ViT and subsequently concatenated for LLM processing. A collection of methods have adopted cropping to scale up resolution~\cite{ureader,lin2023sphinx,li2023monkey,xc2-4k}. 
While effective in reducing ViT complexity, cropping compromises the structural integrity of the image, thus potentially impacting overall performance. 
Moreover, the proliferation of visual tokens introduced by cropping poses significant complexity on LLMs and challenges the retrieval capabilities of LLMs~\cite{xc2-4k}.

\textbf{High-resolution LMMs with Extra Visual Encoders.} 
Incorporating an auxiliary visual encoder for high-resolution image understanding would not significantly increase the number of visual tokens. Vary~\cite{vary} and Deepseek-VL~\cite{deepseek-vl} utilize SAM~\cite{sam} as a high-resolution visual encoder to augment the feature of ViT. MiniGemini-HD~\cite{minigemini} and LLaVA-HR~\cite{llava-hr} employ ConvNeXt~\cite{openclip} to process high-resolution images and use cross-attention or adapters to extract features from the high-resolution input. However, these methods introduce additional complexity through supplementary visual encoders and associated hyperparameters. Furthermore, extracting features from low-quality representations (e.g., LAION-CLIP-ConvNeXt) may potentially compromise LMMs' performance~\cite{gadre2024datacomp,metaclip}.

\section{ConvLLaVA}
\label{sec:method}
We present ConvLLaVA, as illustrated in Fig.~\ref{fig:structure}~(b), whose visual encoder is a five-stage ConvNeXt.
We first introduce the overall architecture and the advantages of our ConvLLaVA in Section~\ref{sec:convllava}.
The two major optimizations: updating the visual encoder and training an additional stage are introduced in Section~\ref{sec:updating} and Section~\ref{sec:add-stage}.

\subsection{ConvNeXt as Standalone Visual Encoder}
\label{sec:convllava}
\begin{figure}
    \centering
    \includegraphics[width=\textwidth]{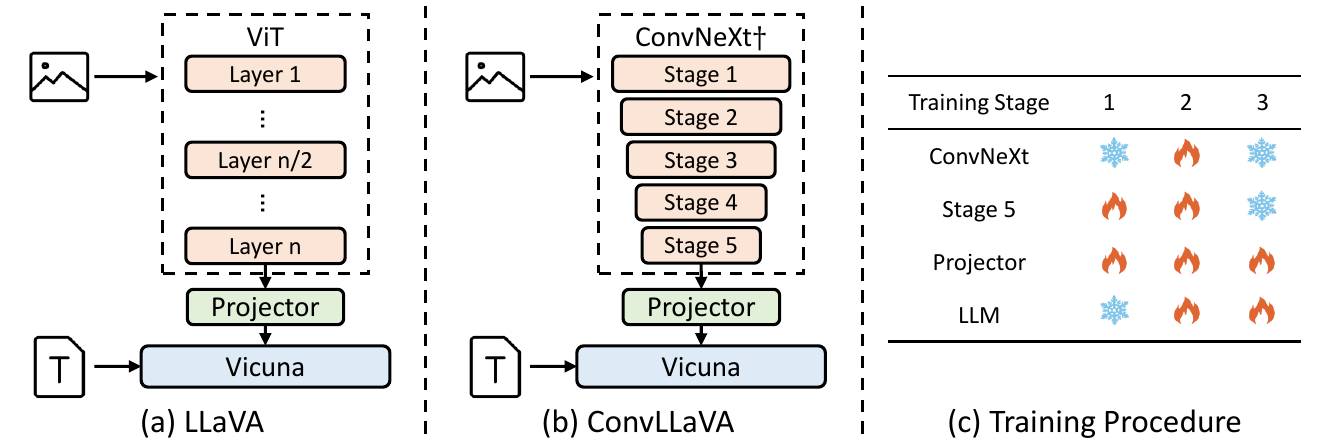}
    \caption{We show the structure for LLaVA and ConvLLaVA in (a) and (b). ConvNeXt has a hierarchical structure which compresses visual tokens between stages. The training procedure is composed of three training stages and the trainable parameters for each stage are shown in (c). }
    \label{fig:structure}
\end{figure}

The architecture of ConvLLaVA is identical to most popular general LMMs, \textit{e.g.}, LLaVA~\cite{llava-v1,llava-v1-5}, Qwen-VL~\cite{qwen-vl}, and VILA~\cite{lin2023vila}. 
These models comprise three components as shown in Fig.~\ref{fig:structure}~(a): a vision encoder $g()$, a large language model $f()$, and a vision-language projector $h()$.
Specifically, the vision model encodes the visual inputs $\vx$ into latent visual embeddings $g(\vx)$. 
The vision-language projector then maps the latent visual embeddings into the embedding space of the language model $\vz = h(g(\vx))$. 
Given the visual embeddings $\vz$ and text embeddings $\vt$ encoded by the language tokenizer, these embeddings are concatenated along the sequence dimension and then passed to the language model. 
Finally, the vision language model is trained with language modeling loss~\cite{gpt}.
Considering that our study mainly focuses on the visual encoder, we employ a two-layer MLP and Vicuna-7B~\cite{vicuna} as the projector and language model following LLaVA-1.5~\cite{llava-v1-5}.
Rather than using CLIP-VIT~\cite{clip}, we introduce CLIP-ConvNeXt~\cite{liu2022convnet,openclip} as the standalone visual encoder.

\begin{wrapfigure}{r}{0.4\linewidth}
    \centering
    \includegraphics[width=\linewidth]{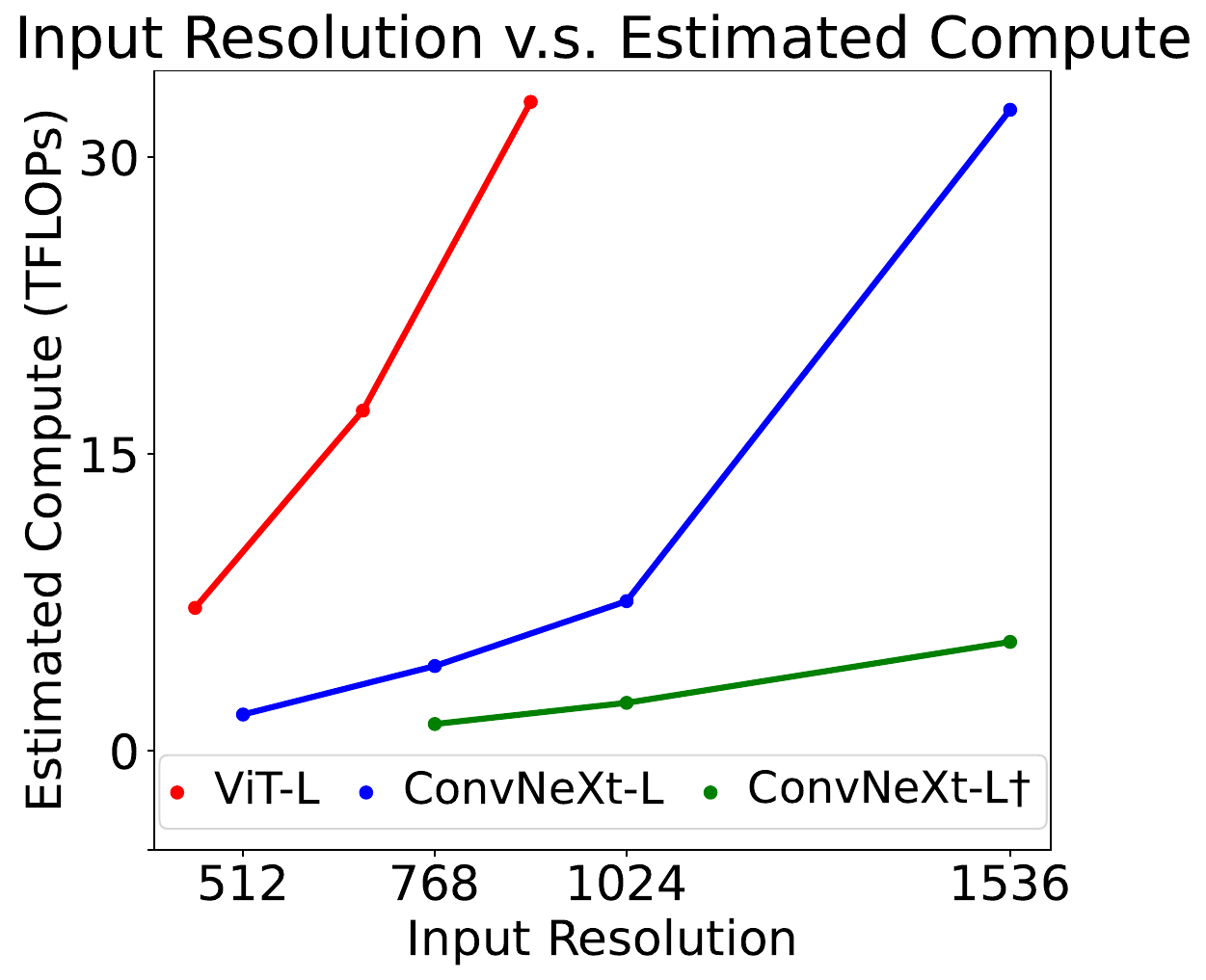}
    \caption{Estimated FLOPs v.s. image resolution for different backbones. We estimate total FLOPs with the number of visual tokens.}
    \label{fig:quality}
\end{wrapfigure}

\textbf{ConvNeXt.} The basic block of ConvNeXt comprises a depth-wise convolution and a feed-forward network~\cite{liu2022convnet}.
The depth-wise convolution has a \textit{7$\times$7} kernel size, and the computation complexity is $\mathcal{O}(k^2CN)$, where $k$, $C$, and $N$ are the kernel size, number of channels, and number of visual tokens, respectively. 
In contrast, the complexity of self-attention in ViT is $\mathcal{O}(4C^2N+2CN^2)$. Consequently, the spatial complexity of ConvNeXt is significantly lower than ViT.
The input is initially processed by a \textit{4$\times$4} non-overlapping convolution downsampling layer.
Subsequently, the features are successively fed into the four stages of ConvNeXt, while each stage comprises several ConvNeXt blocks. Feature maps are downsampled by \textit{2$\times$}, and dimensions are expanded by \textit{2$\times$} between stages.
The output of the ConvNeXt is downsampled by \textit{32$\times$}, rather than \textit{14$\times$} of ViT-L. Hence, ConvNeXt produces less than \textit{1/4} visual tokens compared to ViT, which alleviates the computation load of the language model.
Benefiting from the linear spatial complexity and fewer visual tokens, the computation reduction of LMMs from ViT-L (\textcolor{red}{red} line) to ConvNeXt (\textcolor{blue}{blue} line) is almost \textit{8$\times$} as illustrated in Fig.~\ref{fig:quality}.

\textbf{Five-stage ConvNeXt$\dag$.} Leveraging ConvNeXt as the visual encoder is efficient for encoding 768 resolution images, while scaling resolutions to higher than 768 produces excessive visual tokens.
Previous studies~\cite{llava-v1-6,minigemini} neglect to explore compressing visual tokens, while compressing visual tokens has been proven to be reasonable since there is redundancy in the visual representation~\cite{lin2023vila,fastv}.
These studies suggest that we can further downsample visual features using ConvNeXt.
We propose to compress visual features by incorporating ConvNeXt blocks for stage 5 into the original four-stage model.
We prefer using ConvNeXt blocks over other structures due to the following three reasons
(1) The five-stage ConvNeXt, as a whole, could be transferred as a visual encoder for other LMMs, whereas downsampling in the projector does not offer such flexibility 
(2) ConvNeXt blocks maintain translation equivariance, allowing them to effectively process images of any aspect ratio, unlike attention blocks. 
(3) The impact on performance from the downsampling stage is minimal, except that the resampler consistently underperforms compared to other methods, as evidenced by~\cite{honeybee,xc2-4k,mm1}.
Finally, we denote the overall five-stage ConvNeXt as ConvNeXt$\dag$.
At 1536 resolution, ConvNeXt$\dag$ reduces the number of visual tokens to 576, equivalent to that of ViT at 336 resolution. This would reduce the total computation by \textit{6$\times$} \textit{w.r.t.} the original ConvNeXt (\textcolor{blue}{blue} line) to ConvNeXt$\dag$ (\textcolor{teal}{green} line) as shown in Fig.~\ref{fig:quality}.
Our approach is more computationally efficient than cropping methods, which often produce an excessive number of visual tokens~\cite{mm1,llava-v1-6,li2023monkey}. Furthermore, by eliminating the need for cropping and merging, ConvLLaVA avoids the global view, thereby further reducing the number of visual tokens.

\subsection{Updating ConvNeXt is Essential}
\label{sec:updating}

The mainstream optimization approach~\cite{llava-v1,lin2023vila} freezes the vision encoder during training, as it has better performance and is more efficient than updating the visual encoder~\cite{prismatic}. However, freezing ConvNeXt during training is sub-optimal. 
Hence, we conduct depth analysis to prove that freezing the visual encoder (i.e., ConvNeXt) would inherit the defects from pretraining, and updating ConvNeXt may both improve the quality of representations and adapt them to high-resolution inputs. 

\textbf{Setups of Freezing ConvNeXt.}
The optimization procedure is the same as LLaVA-1.5~\cite{llava-v1-5}. 
For training the projector and instruction tuning, we use the same 558k caption dataset and 665k instruction data, respectively. 
Our visual encoder CLIP-ConvNeXt-L is pretrained on 256 resolution and fine-tuned with 320 resolution based on LAION-2B~\cite{liu2022convnet,openclip}.
We directly increase the resolution to 512 and 768 when applying ConvNeXt as the vision encoder.
As for the baseline, we use ViT which is pretrained on 336 resolution with OpenAI WIT dataset~\cite{clip}.
The training and inference speed for ConvNeXt on 768 resolution is on par with ViT on 336 resolution.
Hence, we consider the comparison between 768-resolution ConvNeXt and 336-resolution ViT to be fair. Detailed training procedure is shown in~Tab.~\ref{tab:hy-llava}.

\textbf{Benchmarks.} 
We use four standard benchmarks to evaluate the results: two general capability benchmarks, MMbench~\cite{liu2023mmbench}, SEEDBench~\cite{li2023seed}, and two fine-grained OCR benchmarks, TextVQA~\cite{textvqa} and DocVQA~\cite{docvqa}. 
It is worth noting that our evaluation procedure for TextVQA differs slightly from LLaVA-1.5~\cite{llava-v1-5}, as we use VLMEVALKIT which does not include OCR tokens in the question. 

\textbf{Results for Freezing the Visual Encoder.}
As shown in Tab.~\ref{tab:freezing-encoder}, we observe the following results:

(1) ConvNeXt has significant advantages over ViT on OCR benchmarks. On TextVQA and DocVQA, both 512 and 768 resolution ConvNeXt outperforms ViT due to their higher resolution~\cite{prismatic,mplug-owl2}. Even with fewer visual tokens, the 512-resolution ConvNeXt still outperforms the 336-resolution ViT.

(2) The overall general capability of ConvNeXt is inferior to ViT. For general benchmarks, on SEEDBench, 768-resolution ConvNeXt performs comparably with ViT. While on MMBench, ConvNeXt underperforms ViT. We hypothesize that there are two reasons for the performance gap on MMbench: First, ConvNeXt is pretrained on low resolution but directly applied on high resolution. Such employment affects the quality of visual features. Second, the pretrained representation for ConvNeXt may be inferior to OpenAI's ViT~\cite{clip}. 

The results imply that increasing resolution without training could affect the quality of representation and hamper the performance of LMMs. However, studies have shown that simply updating the visual encoder during instruction tuning can hinder performance~\cite{prismatic}. To mitigate this issue, ShareGPT4V~\cite{sharegpt4v} provides an effective training protocol and a high-quality dataset for updating the visual encoder. Therefore, we adopt this effective method to update the visual encoder.

\renewcommand{\arraystretch}{1.1}
\begin{table}[!t]
    \caption{Comparison on different visual encoders. Visual encoders are frozen during training. \#Params, Res., PT Data is short for Number of Parameters, Resolution, and Pretraining Data. }
    \label{tab:freezing-encoder}
    \centering
    \scalebox{1}{
    \tabcolsep3pt
    \begin{tabular}{cccccccc}
    \toprule
        Visual Encoder & \#Params & Res. & PT Data & MMBench & SEEDBench &  TextVQA & DocVQA \\ \midrule
        ViT-L & 304M & 336 & WIT & 66.5 & 65.3 & 45.5 & 21.2 \\ \midrule
        ConvNeXt-L & 200M & 512 & LAION-2B & 64.5 & 63.8 & 49.3 & 23.0 \\ 
        ConvNeXt-L & 200M & 768 & LAION-2B & 63.2 & 65.7 & 53.7 & 29.8 \\   \bottomrule 
    \end{tabular}}
\end{table}

\renewcommand{\arraystretch}{1.1}
\begin{table}[!t]
    \caption{Comparison on different visual encoders. The visual encoders are updated during training. The number in the bracket is the improvement compared with freezing the visual encoder. The absolute improvement shown in \textcolor{OliveGreen}{green} is according to the results in Tab.~\ref{tab:freezing-encoder}. \#V Tokens stands for the number of visual tokens.}
    \label{tab:ShareGPT4V}
    \centering
    \scalebox{1}{
    \tabcolsep3pt
    \begin{tabular}{cccccccc}
    \toprule
        Visual Encoder & Res. & \#V Tokens & MMBench & SEEDBench & TextVQA & DocVQA & $\textcolor{OliveGreen}{\Delta}$ \\ \midrule 
        ViT-L & 336 & 576 & \textbf{66.8}\textcolor{OliveGreen}{(+0.3)} & 68.1\textcolor{OliveGreen}{(+2.8)} & 50.1\textcolor{OliveGreen}{(+4.6)} & 26.4\textcolor{OliveGreen}{(+5.2)} & \textcolor{OliveGreen}{+3.2}\\ \midrule
        ConvNeXt-L & 512 & 256 & 65.5\textcolor{OliveGreen}{(+1.0)} & 67.9\textcolor{OliveGreen}{(+4.1)} & 55.1\textcolor{OliveGreen}{(+5.8)} & 31.4\textcolor{OliveGreen}{(+8.4)} & \textcolor{OliveGreen}{+4.8}\\ 
        ConvNeXt-L & 768 & 576 & 66.5\textcolor{OliveGreen}{(+3.3)} & \textbf{68.6}\textcolor{OliveGreen}{(+2.9)} & \textbf{60.0}\textcolor{OliveGreen}{(+6.3)} & \textbf{40.2}\textcolor{OliveGreen}{(+10.4)} & \textbf{\textcolor{OliveGreen}{+5.7}} \\ 
        \bottomrule 
    \end{tabular}}
\end{table}

\textbf{Setups of Updating ConvNeXt.} To update the visual encoder, we first leverage the 558k caption dataset for projector initialization~\cite{llava-v1-5}.
Then, we apply a high-quality caption dataset, ShareGPT4V-PT~\cite{sharegpt4v}, to train the entire vision-language model including the visual encoder.
Finally, the LLaVA 665k instruction tuning dataset is used for visual instruction tuning.
The detailed training procedure is shown in~Tab.~\ref{tab:hy-sharegpt4v}. The last 12 layers of ViT-L are trainable~(according to ShareGPT4V~\cite{sharegpt4v}). For ConvNeXt, we update the last 18 blocks~(ConvNeXt-L  has a total of 36 blocks).

\textbf{Results for Updating the Visual Encoder.}
As shown in Tab.~\ref{tab:ShareGPT4V}, we observe the following results:

(1) ConvNeXt has significant advantages over ViT on the OCR benchmark. The improvement for 768 resolution ConvNeXt is larger than 336 resolution ViT~(6.3/10.4 \textit{v.s.} 4.6/5.2). These results demonstrate the idea of compressing high-resolution visual inputs to a small number~(\textit{e.g.}, 576) of information-rich visual tokens is feasible. Compressing does not lead to great information loss. Even with the same number of tokens, ConvNeXt preserves more fine-grained visual information and significantly outperforms ViT. 

(2) For general benchmarks, ConvNeXt performs on par with ViT. Specifically, ConvNeXt outperforms ViT on SEEDBench and performs on par with ViT on MMBench. Notably, the performance gap between the 768 resolution ConvNeXt and the 336 resolution ViT on MMBench is narrowed from 3.3 to 0.3 compared with freezing the visual encoder. This implies that updating the visual encoder is essential. To further support this, we show the results of updating the visual encoder with more data in Appendix~\ref{app:more-data}. 

Generally, the updated ConvNeXt performs better than ViT on these 4 benchmarks. This evidences that updating the ConvNeXt significantly enhances the performances, underscoring its critical importance. Previous methods employ ConvNeXt as an auxiliary visual encoder and directly increase the resolution to 1024~\cite{llava-hr} or 1536~\cite{minigemini}. They fail to identify the problem that scaling up the resolution without updating ConvNeXt would compromise the performance. Our method, delving deeper into the root of the issue, provides a simple yet effective solution to scaling up the resolution. 

\subsection{Training with Stage 5 Scales up Resolution to 1536}
\label{sec:add-stage}

As we mentioned in Section~\ref{sec:convllava}, scaling resolution to higher than 768 would generate excessive visual tokens. To reduce the redundancy and mitigate the excessive computational demands on the large language model (LLM), we propose training stage 5 for the ConvNeXt model to compress the visual information~(training protocol shown in Fig.~\ref{fig:structure}~(c)).

\renewcommand{\arraystretch}{1.1}
\begin{table}[!t]
\caption{Results of training stage 5 ConvNeXt model. The number of visual tokens does not greatly increase when scaling up resolution. }
\label{tab:add-stage}
    \centering
    \scalebox{0.96}{
    \tabcolsep3pt
    \begin{tabular}{cccccccc}
    \toprule
         Visual Encoder & Resolution & \#Visual Tokens & MMBench & SEEDBench & TextVQA & DocVQA \\ \midrule 
        ConvNeXt-L$\dag$ & 768 & 144 & 65.3 & 67.7 & 54.7 & 31.1 \\ 
        ConvNeXt-L$\dag$ & 1024 & 256 & 65.1 & 68.3 & 59.4 & 35.8 \\ 
        ConvNeXt-L$\dag$ & 1536 & 576 & 64.3 & 69.1 & 60.7 & 42.5 \\    \bottomrule
    \end{tabular}}
\end{table}

\textbf{Implementation Details.} We employ a three-stage training protocol. In the projector initialization stage, we train the fifth stage layers and the projector with the ShareGPT4V-PT data~\cite{sharegpt4v}. In the second stage, we train the entire model with the ShareGPT4V-PT data. For instruction tuning, we utilize the 665k LLaVA instruction data to train the LLM and the projector. The training protocol is similar to the protocol for updating the visual encoder. The only difference is that we train the fifth stage and projector with ShareGPT4V-PT data, while experiments in Section~\ref{sec:updating} train the projector with the 558k caption data in the first training stage. We add 6 layers in stage 5 and tune the last three stages in the second training phase. Ablation studies on these hyper-parameters are included in Appendix~\ref{app:stage-add-layers}.

\textbf{Results for ConvNeXt$\dag$.} We present the results of adding stage 5 to ConvNeXt in Tab.~\ref{tab:add-stage}. Scaling up the resolution consistently improves performance on SEEDBench, TextVQA, and DocVQA, which require fine-grained understanding and benefit from the higher resolution. These results highlight the effectiveness of our method of training stage 5. However, on MMBench, the performance of ConvNeXt$\dag$ exhibits a slight drop when scaling the resolution from 1024 to 1536. The resolution of 1536 is approximately six times higher than the pretraining resolution~(256). Adapting the pretrained visual encoder to effectively extract global information from such a significant increase in resolution requires a substantial amount of training data. In Section~\ref{sec:exp}, we verify this hypothesis by providing sufficient data to the visual encoder in the second training stage.

\begin{figure}[t]
  \centering
  \begin{subfigure}[b]{0.49\linewidth}
    \flushright
    \includegraphics[width=\linewidth]{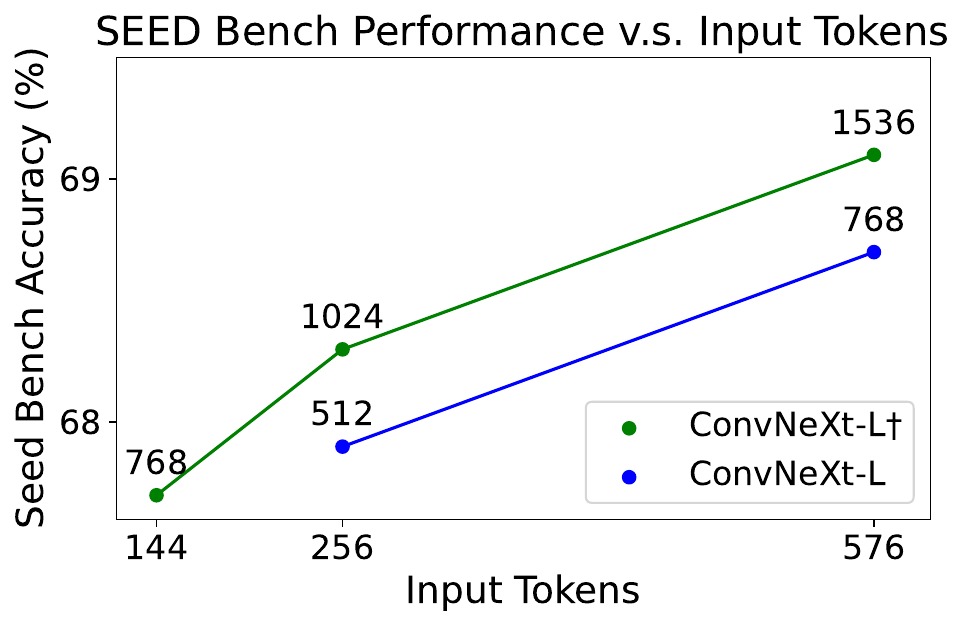}
  \end{subfigure}
  \hfill
  \begin{subfigure}[b]{0.49\linewidth}
    \flushleft
    \includegraphics[width=0.99\linewidth]{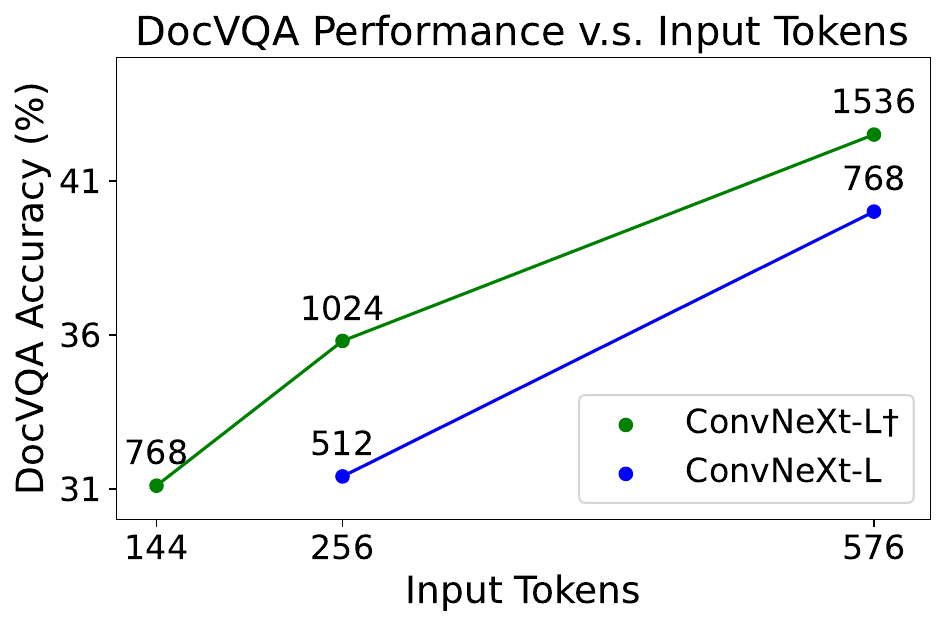}
  \end{subfigure}
  \caption{Comparisons of ConvNeXt and ConvNeXt$\dag$ on SEEDBench and DocVQA. The marked number above the line shows the resolution of the model. }
  \label{fig:resolution-tokens}
\end{figure}

\textbf{On Scaling Resolution.} When we increase the resolution, the number of visual tokens also increases. These two factors are entangled, and there has been a lack of in-depth investigation into the relationship between them. Previous work claims that raw resolution matters more than the number of visual tokens~\cite{lin2023vila}. We experiment on the general benchmark SEEDBench and OCR benchmark DocVQA to investigate these assumptions. Our method provides control experiments to reveal the relationship between resolution and the number of visual tokens. We compare the results of ConvNeXt (trained in Section~\ref{sec:updating}) and ConvNeXt$\dag$ (trained in Section~\ref{sec:add-stage}) as the visual encoder for LMMs under the same number of visual tokens. The two series of models are pretrained with ShareGPT4V-PT and instruction-tuned with 665k LLaVA instruction data. ConvNeXt$\dag$ has an additional stage to compress the number of visual tokens to 1/4. Hence, the differences between these two series models have been largely reduced. Our control experiments reveal novel findings:

(1) When the number of visual tokens is the same, the higher resolution model exhibits better performance on SEEDBench and DocVQA. In the Fig.\ref{fig:resolution-tokens}, the green line consistently outperforms the blue line. This is because that high-resolution model provides finer-grained and higher-quality visual features even if the output number of visual tokens is the same. Previous work~\cite{llava-v1-6,li2023monkey,xc2-4k} which scales up the resolution by splitting the image into patches would generate excessive visual tokens. Such cropping methods significantly sacrifice efficiency and challenge the retrieval capability of LLM. Our core discovery presents a promising approach to enrich the information contained in visual features without compromising efficiency. Compressing high-resolution images into information-rich visual tokens is more efficient than the cropping method. Training a stage to further compress visual features provides a manner to increase resolution and maintain a moderate computational cost.

(2) The importance of the number of visual tokens varies across different benchmarks at equivalent resolution. For general benchmarks like SEEDBench, the performance drop brought by compressing visual tokens for the 768-resolution models is marginal (0.9 on SEEDBench). However, for OCR benchmarks like DocVQA, the performance drop for the model with fewer visual tokens is substantial (9.1 on DocVQA). Overall, these results demonstrate that while compressing visual tokens causes only slight information loss on general benchmarks, but leads to significant information loss on fine-grained OCR benchmarks.

\section{Experiments}
\label{sec:exp}

Our results demonstrate that scaling up the resolution of ConvNeXt and updating the visual encoder are two effective approaches to training an advanced, high-resolution Language-Multimodal Model. However, we found that the available training data was insufficient to fully unleash the potential of these approaches. Consequently, we scaled up the high-quality training data to address this limitation. 

\subsection{Training Setups}

\textbf{Training Stages.} We adopt a three-stage training protocol to train ConvLLaVA as shown in Fig.~\ref{fig:structure}~(c).
The training process is categorized into three stages:
(1) \textit{Projector Initialization.} We train the fifth stage of the ConvNeXt model and the vision-language projector. We utilize caption data including ShareGPT4V-PT~\cite{sharegpt4v}, ShareGPT4V~\cite{sharegpt4v}, and ALLaVA captions~\cite{allava}, totaling approximately 2M examples. 
(2) \textit{Vision-Language Pretraining.} We employ caption data including ShareGPT4V-PT~\cite{sharegpt4v}, ShareGPT4V~\cite{sharegpt4v}, ALLaVA~\cite{allava}, and a 190k open-sourced subset of VFLAN~\cite{vflan}, amounting to 2.9M data. 
(3) \textit{Visual Instruction Tuning.} We fine-tune the model with the 665k LLaVA instruction dataset~\cite{llava-v1-5}. In each stage, we train the model for 1 epoch with the AdamW optimizer. The cosine learning rate schedule is also applied. 

\textbf{Implementation Details.} We utilize the LAION-2B pretrained ConvNeXt-L model as our visual encoder~\cite{openclip}. In the three training stages, the resolution is scaled up to a fixed value. We train ConvLLaVA at 768, 1024, and 1536 resolutions. 
The learning rates in the three training stages are 3e-4, 2e-5, and 2e-5, respectively. Meanwhile, the batch sizes are 256, 256, and 128.
Training the ConvLLaVA 768 resolution model takes approximately 18 hours on 2 A800 machines. The instruction tuning costs 20 hours for LLaVA-NExT 7B on an A100 machine~\cite{llava-v1-6}, while it tasks only 9 hours for our 1536 resolution ConvLLaVA on a single machine.

\textbf{Evaluation Benchmarks.} To systematically investigate the performance of our model, we include more benchmarks for evaluation, including MME~\cite{mme}, MMBench~\cite{liu2023mmbench}, SEEDBench~\cite{li2023seed}, MMMU~\cite{yue2023mmmu}, MMVet~\cite{mmvet}, RealWorldQA~\cite{grok1_5}, TextVQA~\cite{textvqa}, DocVQA~\cite{docvqa}, and POPE~\cite{pope}. Our results are measured by VLMEVALKIT. We also assess the performance on grounding benchmarks, including RefCOCO~\cite{refcoco}, RefCOCO+, and RefCOCOg~\cite{refcocog}.

\renewcommand{\arraystretch}{1.1}
\begin{table}[t!]
\caption{Comparisons with different resolution multi-modality models. $^*$The results are measured by VLMEVALKIT with official checkpoints. $^\dag$The results are measured with the original image aspect ratio and the short side of the image is resized to 1536. $^\ddag$OtterHD is tested with the original resolution of the images and the number of tokens varies. We mark the best performance of the 7B model \textbf{bold} and the second-best \underline{underlined}.}\vspace{-5pt}
\label{tab:main}
\flushleft
\scalebox{0.9}{
\tabcolsep2pt
\begin{tabular}{ccccccccccccc}
\toprule
Method        & Res. & \#V Tokens & LLM & MME  & MMB  & SEED & RWQA & MMMU & MMVet & Text & Doc  & POPE \\ \midrule
LLaVA-1.5     & 336        & 576           & 13B & 1531 & 68.2 & 68.2 & 55.3        & 36.4 & 38.3  & 48.7$^*$ & 23.7$^*$ & 85.9 \\
VILA          & 336        & 576           & 13B & 1570 & 70.3 & --     &    --         &  --    & 38.8  & 54.5$^*$ & 39.2$^*$ & 84.2 \\
LLaVA-Next    & 672        & 2880          & 13B & 1575 & 70   & 71.9 &      --       & 36.2 & 48.4  & 67.1$^*$ & -- & 86.7 \\ \midrule
OtterHD       &   --$^\ddag$         &    --$^\ddag$           & 8B  & 1223 & 58.3 &   --   &        --     &    --  & 26.3  & --     &   --   & 86   \\
Qwen-VL-Chat  & 448        & 256           & 7B  & 1488 & 60.6 & 64.8 &     --        & --   & --  &  61.5  & \underline{62.6} &    --  \\
LLaVA-1.5     & 336        & 576           & 7B  & 1510 & 64.3 & 66.2 & 54.8        & -- & 30.5  & 45.5$^*$ &  21.6$^*$    & 85.9 \\
ShareGPT4V    & 336        & 576           & 7B  & \underline{1567} & \underline{68.8} & \underline{69.7} & 56.5        &  --    & 37.6  & 51.1$^*$ & 26.6$^*$ &  86    \\
VILA          & 336        & 576           & 7B  & 1533 & \textbf{68.9} &   --   &    --         &  --    & 35.1  & 53.2$^*$ & 35.8$^*$ & 86.3 \\
LLaVA-Next    & 672        & 2880          & 7B  & 1519 & 67.4 & \textbf{70.2} &      --       & 35.8 & 43.9  & \underline{64.4}$^*$ & -- & 86.5 \\
MiniGemini-HD & 1536       & 2880          & 7B  & 1546 & 65.8 &    --  &     --        &\textbf{36.8} & 41.3  & --    &    --  &    --  \\
\midrule 
\rowcolor{cyan!15}ConvLLaVA     & 768        & 144           & 7B  & 1541 & 68   & 68.8 & 55.9        & \underline{36.3} & \underline{44.8}  & 59.1 & 44.8 & \underline{87.3} \\
\rowcolor{cyan!15}ConvLLaVA     & 1024       & 256           & 7B  & 1553 & \underline{68.8} & 69.3 & \underline{58.8}        & 35.1 & 44.4  & 62.5 & 48.5 & \textbf{87.7} \\
\rowcolor{cyan!15}ConvLLaVA     & 1536       & 576           & 7B  & \textbf{1575} & 68.7 & \textbf{70.2} & \textbf{59.9}        & 35.8 & \textbf{45.9}  & \textbf{65.8} & 59/\textbf{65}   & \underline{87.3} \\  
\bottomrule
\end{tabular}}
\end{table}

\renewcommand{\arraystretch}{1.1}
\begin{table}[t!]
\caption{Results on referring expression comprehension tasks. The models in this table are trained with the same grounding data. We mark the best performance of the model \textbf{bold}.}
\label{tab:grounding}
\centering
\tabcolsep2pt
\begin{tabular}{ccccccccccccc}
\toprule
\multirow{2}{*}{Method} & \multirow{2}{*}{Res.} & \multirow{2}{*}{\#V Tokens} & \multirow{2}{*}{LLM} & \multicolumn{3}{c}{RefCOCO} & \multicolumn{3}{c}{RefCOCO+} & \multicolumn{2}{c}{RefCOCOg} & \multirow{2}{*}{Avg} \\ 
                        &                             &                                &                      & val    & test-A   & test-B  & val    & test-A   & test-B   & val           & test         &                      \\           \midrule
LLaVA-1.5               & 336                         & 576                            & 7B                  & 76.3 & 83.2 & 67.9 & 66.8 & 77.0 & 56.8 & 70.4 & 70.0 & 71.1                \\
LLaVA-1.5               & 336                         & 576                            & 13B                  & 84.0   & 89.5     & 77.1    & 76.3   & 84.3     & 66.1     & 78.8          & 78.3         & 79.3                 \\ \midrule
\rowcolor{cyan!15}ConvLLaVA               & 768                         & 144                            & 7B                   & 84.5   & 89.0     & 79.2    & 77.7   & 84.9     & 69.7     & 79.8          & 79.7         & 80.6                 \\
\rowcolor{cyan!15}ConvLLaVA               & 1024                        & 256                            & 7B                   & 85.5   & 89.6     & 78.8    & 79.3   & 86.1     & 70.3     & 80.6          & 81.2         & 81.4                 \\
\rowcolor{cyan!15}ConvLLaVA               & 1536                        & 576                            & 7B                   & \textbf{86.5}   & \textbf{90.6}     & \textbf{80.5}    & \textbf{80.0}   & \textbf{86.8}     & \textbf{71.5}     & \textbf{82.0}          & \textbf{82.4}         & \textbf{82.3}      \\ \bottomrule          
\end{tabular}
\end{table}

\subsection{Quantitative Results}

We perform a comprehensive comparison with state-of-the-art models on 7 different benchmarks~(Tab.~\ref{tab:main}). Our model achieves consistent improvements compared to LLaVA-1.5. Our 7B model even exhibits comparable performance with LLaVA-1.5 13B and LLaVA-NExT 7B~\cite{llava-v1-6}. On OCR benchmarks like TextVQA and DocVQA, our model outperforms the LLaVA-1.5 7B and 13B models. Since OCR benchmarks are sensitive to resolution, our ConvLLaVA series models demonstrate consistent improvement on TextVQA and DocVQA with higher resolution, showcasing the effectiveness of scaling up resolution. Notably, our model surpasses Qwen-VL-Chat on DocVQA which has millions of document training data. While there is only a limited number of document data in our training dataset. This shows the benefits of the high-resolution design of our model. ConvLLaVA outperforms LLaVA-NExT on MMBench, TextVQA, POPE, and MMVet. 

For grounding benchmarks, our model and LLaVA are trained with the same set of grounding data. The comparison between them is fair. On RefCOCO, RefCOCO+, and RefCOCOg, ConvLLaVA exhibits consistent improvement when increasing resolution~(Tab.~\ref{tab:grounding}). ConvLLaVA outperforms LLaVA-7B and 13B model on all 8 test splits. This demonstrates the benefits of higher resolution for grounding tasks. Our 7B model also surpasses 13B LLaVA model on all 8 benchmarks. 

\subsection{Understanding Any Aspect Ratio Images and Highre Resolutions}

Thanks to the translation equivalence of convolution neural network, our model could be trained on a fixed resolution but inference on higher resolution and with an arbitrary aspect ratio. We test such ability on our 1536 resolution model ConvLLaVA. 

\begin{wraptable}{r}{0.4\linewidth}
\caption{Results for different aspect ratios and higher resolution. }\vspace{-5pt}
    \tabcolsep3pt
    \centering
    \begin{tabular}{c|ccc}
    \toprule
       Input Shape & SEED & Text & Doc \\ \midrule
        (1536, 1536) & \textbf{70.2} & \textbf{65.8} & 59.0 \\ 
        short side=1536   & 68.9 & 64.6 & 65.0 \\ 
        short side=1664   & 67.3 & 64.2 & \textbf{65.7} \\ 
        \bottomrule
    \end{tabular}
    \label{tab:shape}
\end{wraptable}

The original image preprocessing process is padding the image to a square, resizing the image to 1536, and center cropping~\cite{llava-v1-5}. We canceling padding and center cropping. Hence, the short side of the image should just be resized to 1536 and keep the original aspect ratio. This is the setting of how we test images of any aspect ratio. The results are shown in Tab.~\ref{tab:shape}. We observe that on the general benchmark, SEEDBench, the performance slightly decreases. On OCR benchmarks, especially on DocVQA, the performance is improved. The reason for this we think is that the image aspect ratio in DocVQA is not 1:1, forcely transforming the image into a square would lower the resolution of the image. 

We also test ConvLLaVA when resizing the short side of images to 1664 resolution which is higher than its pretrained 1536 resolution. We observe that on DocVQA the performance could be further improved to 65.7.

\subsection{Discussions}
\label{sec:discussions}
\textbf{Architectures and data.} While we have demonstrated the effectiveness of our method, there remains room for further improvement. The ConvNeXt architecture we use is tailored for low-resolution image understanding (e.g., 256), with a kernel size of 7 optimized for such resolutions. However, as the resolution increases to 1536, the relatively small kernel size may limit the model capacity when the resolution is extremely high. Besides, the number of layers in the ConvNeXt four stages (3, 3, 27, 3) is designed for a 4-stage model and may not be optimal for our 5-stage model. Therefore, a potential future direction could involve designing a five-stage, linear spatial complexity, hierarchical high-resolution vision encoder. We emphasize the critical role of the five-stage visual encoder since it is fit for high-resolution LMM. It compresses visual features by \textit{64$\times$}, greatly reducing the redundancy in its visual tokens. In contrast, four-stage visual encoders, designed for traditional computer vision tasks, output excessive tokens when resolution is high.

\textbf{Linear spatial complexity and information compression. }We identify \textit{linear spatial complexity} and \textit{information compression} procedure as two critical properties for future visual encoders of LMMs. These properties ensure the efficiency of both the visual encoder and the LLM, respectively. Furthermore, they are crucial for multi-image, interleaved image and text, and video understanding tasks, as these tasks commonly result in numerous visual tokens. We anticipate that future research will focus more on these two directions to further advance the research of LMMs.

\textbf{Trade-off between compression and retrieval for high-resolution understanding. }Our method, ConvLLaVA, compresses a 1536-resolution image to 576 visual tokens with a 64$\times$ compression ratio. While concurrent work~\cite{xc2-4k,internvl1.5} explores retrieving fine-grained image information from long visual token sequences. In the context of high-resolution image understanding, compressing visual information maintains computational efficiency, but excessive compression may lead to information loss. Conversely, retaining a large number of visual tokens avoids information loss but sacrifices efficiency and challenges the retrieval capabilities of LLMs. Consequently, a trade-off emerges between visual information compression and retrieval capabilities for high-resolution understanding. Future research should explore an optimal balance between these two factors.

\section{Conclusion}

In this paper, we have critically examined the limitations of the visual encoder for current LMMs: quadratic spatial complexity and numerous visual tokens. The excessive visual tokens are the more fundamental problem. These drawbacks hinder LMMs from efficiently understanding high-resolution images. Consequently, we propose ConvLLaVA, whose visual encoder is a hierarchical backbone, ConvNeXt, to mitigate this issue. ConvLLaVA compresses high-resolution visual information into information-rich visual representation rather than preserving all the redundancy in the visual representation. Extensive experimental results have demonstrated the efficacy of our proposed method. Our 7B parameter model exhibits superior performance compared to the LLaVA-1.5 13B model. Furthermore, our method is flexible in encoding images with arbitrary shapes and resolutions. Our work highlights the advantages of hierarchical visual backbones for LMMs, addressing critical challenges while maintaining simplicity and efficiency.

\section*{Acknowledgments}

This work is supported in part by the National Natural Science Foundation of China under Grants 62321005 and 62276150.

\clearpage
\bibliography{ref}
\bibliographystyle{plain}

\clearpage
\appendix
\appendix

\section{Training Visual Encoder with More Data}
\label{app:more-data}
In Section~\ref{sec:updating}, we observe that updating the visual encoder is essential for ConvNeXt as the standalone encoder. We compare the two visual encoders with more training data in Tab.~\ref{tab:allava-sharegpt4v}. For the visual language training stage, we use ALLaVA and ShareGPT4V-PT. We train the last two stages for ConvNeXt and the last 12 layers for ViT. With more training data, ConvNeXt outperforms ViT on all the 4 benchmarks. These results validate the advantages of ConvNeXt over ViT. This ConvNeXt model even outperforms the 768-resolution ConvLLaVA model on some benchmarks due to its higher number of visual tokens. However, the training and inference speed is much slower than the 768-resolution ConvLLaVA model due to the increased number of visual tokens. The 1536 resolution ConvLLaVA, featuring outputting the same number of visual tokens, outperforms this model. This shows higher resolution model may have a higher model capacity to learn from data.

\renewcommand{\arraystretch}{1.1}
\begin{table}[!h]
    \caption{Comparison on different visual encoders. The visual encoders are updated during training. \#V Tokens stands for the number of visual tokens. }
    \label{tab:allava-sharegpt4v}
    \centering
    \scalebox{1}{
    \tabcolsep3pt
    \begin{tabular}{ccccccc}
    \toprule
        Visual Encoder & Res. & \#V Tokens & MMBench & SEEDBench & TextVQA & DocVQA  \\ \midrule 
        ViT-L & 336 & 576 & 68.1 & 68.0 & 49.6 & 29.2 \\ \midrule
        ConvNeXt-L & 768 & 576 & \textbf{68.5} & \textbf{69.5} & \textbf{62.5} & \textbf{51.6} \\ 
        \bottomrule 
    \end{tabular}}
\end{table}

\section{Hyperparameters for 5-stage ConvNeXt}
\label{app:stage-add-layers}

We discuss the choice of hyperparameters in this section. 

\renewcommand{\arraystretch}{1.1}
\begin{table}[!t]
\caption{Ablation on the number of trainable stages. }
\label{tab:stages-high}
    \centering
    \scalebox{1}{
    \tabcolsep3pt
    \begin{tabular}{ccccccc}
    \toprule
        Visual Encoder & Tune from Stage & MMBench & SEEDBench & TextVQA & DocVQA \\ \midrule
        ConvNeXt-L$\dag$ & 2 & 65.1 & 67.7 & 54.8 & 31.1 \\ 
        \rowcolor{cyan!15}ConvNeXt-L$\dag$ & 3 & 65.3 & 67.7 & 54.7 & 31.1 \\ 
        ConvNeXt-L$\dag$ & 4 & 66.2 & 67.0 & 52.2 & 28.2 \\ \bottomrule
    \end{tabular}}
\end{table}

\textbf{Number of Trained Stages.} We conduct an ablation study to determine the optimal number of stages for vision-language pretraining at 768 resolution. We find that fine-tuning from stage 3 yields better results than fine-tuning from stage 4 (Tab.~\ref{tab:stages-high}). While the performances of fine-tuning from stage 2 and stage 3 are comparable, we opt for fine-tuning from stage 3 due to its fewer trainable parameters.

\renewcommand{\arraystretch}{1.1}
\begin{table}[!t]
\caption{Ablation on number of layers in stage 5.}
\label{tab:ablation-layers}
    \centering
    \scalebox{1}{
    \tabcolsep3pt
    \begin{tabular}{ccccccc}
    \toprule
         Visual Encoder & \#Layers Added & MMBench & SEEDBench & TextVQA & DocVQA \\ \midrule
        ConvNeXt-L$\dag$ & 3 & 65.2 & 67.9 & 55.6 & 29.6 \\ 
        \rowcolor{cyan!15}ConvNeXt-L$\dag$ & 6 & 65.3 & 67.7 & 54.7 & 31.1 \\ 
        ConvNeXt-L$\dag$ & 9 & 64.6 & 67.9 & 54.6 & 30.1 \\  \bottomrule
    \end{tabular}}
\end{table}

\textbf{Number of Layers in Stage 5.} We ablate on the number of ConvNeXt layers in stage 5. Given that the number of layers in each stage is a multiple of 3 in ConvNeXt-L, we experiment with 3, 6, and 9 layers in stage 5. For simplicity, we perform the experiments on ConvNeXt 768. We observe a slight decrease in performance when adding 9 layers in stage 5 (Tab.~\ref{tab:ablation-layers}). However, it's hard to determine whether adding 3 or 6 layers is more beneficial for these four benchmarks. Hence, we conduct experiment on the 1536 resolution to further investigate this hyperparameter~(Tab.~\ref{tab:add-layers-1536}). The results show that adding 6 layers could be better. We opt for 6 layers in our experiments. 

\renewcommand{\arraystretch}{1.1}
\begin{table}[!ht]
\caption{Experiments on the number of layers in stage 5 on 1536 resolution. }
\label{tab:add-layers-1536}
    \centering
    \scalebox{1}{
    \tabcolsep2pt
    \begin{tabular}{cccccc}
    \toprule
        Visual Encoder & \#Layers in Stage 5 & MMBench & SEEDBench & TextVQA & DocVQA \\ \midrule
        ConvNeXt-L$\dag$  & 3 & 64.6 & 68.4 & 60.6 & 38.8 \\ 
        \rowcolor{cyan!15}ConvNeXt-L$\dag$  & 6 & 64.3 & 69.1 & 60.7 & 42.5 \\  \bottomrule
    \end{tabular}}
\end{table}

\section{Training protocol for each experiment}
\label{app:implementations}
The detailed training hyper-parameters are shown in the following tables.

\begin{table}[ht]
\caption{The training protocol for Tab.~\ref{tab:freezing-encoder}.}
\centering
\begin{tabular}{ccc}
\toprule
Training Stage  & 1              & 2              \\ \midrule
Visual Encoder  &                &                \\
Projector       & \ding{51}            & \ding{51}            \\
LLM             &                & \ding{51}            \\ \midrule
data            & LLaVA LCS-558K & LLaVA SFT 665k \\
lr              & 1e-3           & 2e-5           \\
batch size      & 256            & 128            \\
lr schedule     & cosine decay   & cosine decay   \\
lr warmup ratio & 0.03           & 0.03           \\
epoch           & 1              & 1              \\
optimizer       & AdamW          & AdamW      \\ \bottomrule   
\end{tabular}
\label{tab:hy-llava}
\end{table}

\begin{table}[h]
\caption{The training protocol for Tab.~\ref{tab:ShareGPT4V}. }
\centering
\begin{tabular}{cccc}
\toprule
Training Stage  & 1              & 2             & 3              \\ \midrule
Visual Encoder  &                & \ding{51}           &                \\
Projector       & \ding{51}            & \ding{51}           & \ding{51}            \\ 
LLM             &                & \ding{51}           & \ding{51}            \\ \midrule
data  & LLaVA LCS-558K & ShareGPT4V-PT & LLaVA SFT 665k 
\\
lr              & 1e-3           & 2e-5          & 2e-5           \\
batch size      & 256            & 256           & 128            \\
lr schedule     & cosine decay   & cosine decay  & cosine decay   \\
lr warmup ratio & 0.03           & 0.03          & 0.03           \\
epoch           & 1              & 1             & 1              \\
optimizer       & AdamW          & AdamW         & AdamW         \\ \bottomrule
\end{tabular}
\label{tab:hy-sharegpt4v}
\end{table}

\begin{table}[h]
\caption{The training protocol for Tab.~\ref{tab:add-stage}, Tab.~\ref{tab:stages-high}, and Tab.~\ref{tab:ablation-layers}}
\centering
\begin{tabular}{cccc}
\toprule
Training Stage  & 1             & 2             & 3              \\ \midrule
ConvNeXt        &               & \ding{51}           &                \\
Stage 5         & \ding{51}           & \ding{51}           &                \\
Projector       & \ding{51}           & \ding{51}           & \ding{51}            \\ 
LLM             &               & \ding{51}           & \ding{51}            \\ \midrule
data            & ShareGPT4V-PT & ShareGPT4V-PT & LLaVA SFT 665k \\
lr              & 3e-4          & 2e-5          & 2e-5           \\
batch size      & 256           & 256           & 128            \\
lr schedule     & cosine decay  & cosine decay  & cosine decay   \\
lr warmup ratio & 0.03          & 0.03          & 0.03           \\
epoch           & 1             & 1             & 1              \\
optimizer       & AdamW         & AdamW         & AdamW         \\ \bottomrule
\end{tabular}
\label{tab:hy-5stages}
\end{table}

\begin{table}[h]
\caption{The training protocol for Tab.~\ref{tab:main}, and Tab.~\ref{tab:grounding}}
\centering
\begin{tabular}{cccc}
\toprule
Training Stage  & 1             & 2             & 3              \\ \midrule
ConvNeXt        &               & \ding{51}           &                \\
Stage 5         & \ding{51}           & \ding{51}           &                \\
Projector       & \ding{51}           & \ding{51}           & \ding{51}            \\ 
LLM             &               & \ding{51}           & \ding{51}            \\ \midrule
\multirow{3}{*}{data}            & ShareGPT4V-PT & ShareGPT4V-PT & \multirow{3}{*}{LLaVA SFT 665k} \\
& ShareGPT4V  & ShareGPT4V  & \\
& ALLaVA Caption & ALLaVA, VFLAN & \\
lr              & 3e-4          & 2e-5          & 2e-5           \\
batch size      & 256           & 256           & 128            \\
lr schedule     & cosine decay  & cosine decay  & cosine decay   \\
lr warmup ratio & 0.03          & 0.03          & 0.03           \\
epoch           & 1             & 1             & 1              \\
optimizer       & AdamW         & AdamW         & AdamW         \\ \bottomrule
\end{tabular}
\label{tab:hy-main-results}
\end{table}

\clearpage

\end{document}